\documentclass{svproc}
\usepackage{cite}
\usepackage{url}
\usepackage{amsmath,amssymb,amsfonts}
\usepackage{algorithmic}
\usepackage{graphicx,color}
\usepackage{textcomp}
\usepackage{xcolor}
\usepackage{algorithm,algorithmic}
\def\BibTeX{{\rm B\kern-.05em{\sc i\kern-.025em b}\kern-.08em
    T\kern-.1667em\lower.7ex\hbox{E}\kern-.125emX}}
\AtBeginDocument{\definecolor{tmlcncolor}{cmyk}{0.93,0.59,0.15,0.02}\definecolor{NavyBlue}{RGB}{0,86,125}}
\usepackage{multirow}
\usepackage{dblfloatfix}
\usepackage[utf8]{inputenc}

\usepackage{listings}
\usepackage{xcolor}

\usepackage{duckuments}
\usepackage{todonotes}
\usepackage{marginnote}

\lstdefinestyle{folderstyle}{
    basicstyle=\ttfamily,
    backgroundcolor=\color{gray!10}
}

\usepackage[nameinlink,capitalise]{cleveref}

\crefname{chapter}{Chap.}{Chaps.}
\Crefname{chapter}{Chapter}{Chapters}
\crefname{section}{Sect.}{Sects.}
\Crefname{section}{Section}{Sections}
\crefname{subsection}{Sect.}{Sects.}
\Crefname{subsection}{Section}{Sections}
\crefname{figure}{Fig.}{Figs.}
\Crefname{figure}{Figure}{Figures}
\crefname{subfigure}{Fig.}{Figs.}
\Crefname{subfigure}{Figure}{Figures}
\crefname{page}{p.}{pp.}
\Crefname{page}{Page}{Pages}
\crefname{equation}{Eq.}{Eqs.}
\Crefname{equation}{Equation}{Equations}
\crefname{table}{Table}{Tables}
\Crefname{table}{Table}{Tables}

\makeatletter
\let\MYcaption\@makecaption
\makeatother

\usepackage{subcaption}

\makeatletter
\let\@makecaption\MYcaption
\makeatother

\usepackage{enumitem}
\usepackage{crossreftools}

\usepackage{graphicx}    
\usepackage{booktabs}    
\usepackage{amssymb} 
\usepackage{gensymb}

\usepackage{siunitx}
\DeclareSIUnit{\byte}{B}

\usepackage{xcolor}

\usepackage[acronym]{glossaries}
\newacronym{vru}{VRU}{Vulnerable Road User}
\newacronym{imu}{IMU}{Inertial Measurement Unit}
\newacronym{av}{AV}{Autonomous Vehicle}
\newacronym{hdmap}{HD Map}{High-Definition Map}
\newacronym{ad}{AD}{Autonomous Driving}
\newacronym{ttc}{TTC}{Time to Collision}
\newacronym{mlp}{MLP}{Multi-Layer Perceptron}
\newacronym{dnn}{DNN}{Deep Neural Network}
\newacronym{mse}{MSE}{Mean Squared Error}
\newacronym{poe}{PoE}{Power over Ethernet}
\newacronym{pbvs}{PBVS}{position-based visual servoing}
\newacronym{ibvs}{IBVS}{image-based visual servoing}
\newacronym{fov}{FOV}{field of view}
\newacronym{mapf}{MAPF}{multi-agent path finding}
\newacronym{cbs}{CBS}{conflict-based search}

\begin{document}

\title{Multi-Robot Planning and Control \\from CCTV Camera Networks in a \\Real Warehouse}
\titlerunning{Multi-Robot Control from CCTV Camera Networks}

\author{Luke Robinson\inst{1} \and Benjamin Ramtoula\inst{1} \and Anas Izaaryene\inst{2} \and Paul Newman\inst{1} \and Daniele De Martini\inst{1}}
\authorrunning{Luke Robinson et al.}
\institute{Oxford Robotics Institute, University of Oxford, UK, \email{lrobinson@robots.ox.ac.uk} \and Robot Systems Group, Technical University of Munich, Germany}

\maketitle


\begin{abstract}
    Off-board control of mobile robots from cameras embedded in the environment offers a practical path to scalable autonomy, moving sensing and compute off the robots.
    We extend this idea from the single-robot case to coordinated fleets in a real warehouse, driving multiple robots with only a distributed CCTV network and edge compute.
    The system operates entirely in image space over an uncalibrated, pixel-wise topological camera graph, enabling wide-area operation with flexible camera placement.
    A hierarchical planner selects a camera sequence per robot and plans its image-space motion through each view, coordinating robots with a prioritised-then-joint strategy and treating overlapping camera regions as shared resources held by one robot at a time to prevent collisions and deadlocks.
    We validate the approach in a real warehouse with four robots and 30 cameras across six \qty{27}{\meter} aisles, reporting mission times and coordination statistics.
    To our knowledge, this is the first field demonstration of multi-robot planning and coordination using only an external camera network and off-board compute, with robots carrying no task-specific navigation hardware.
\end{abstract}

\section{Introduction}

Off-board control of mobile robots from cameras embedded in the environment offers a practical path to scalable autonomy.
With this paradigm, custom, expensive sensors and compute usually carried by each robot are instead placed in the surrounding infrastructure, as edge servers and CCTV-style cameras \cite{buoso2024select2plan,robinson2023iros,robinson2025auro}.
Moving perception, planning, and control off the robots lets these components scale independently of the fleet size \cite{simoens2018internet}.
This is especially attractive in industrial settings such as warehouses, where robots operate in a finite space and the marginal cost per robot drives deployment.

\begin{figure*}
    \centering
    \includegraphics[width=\linewidth]{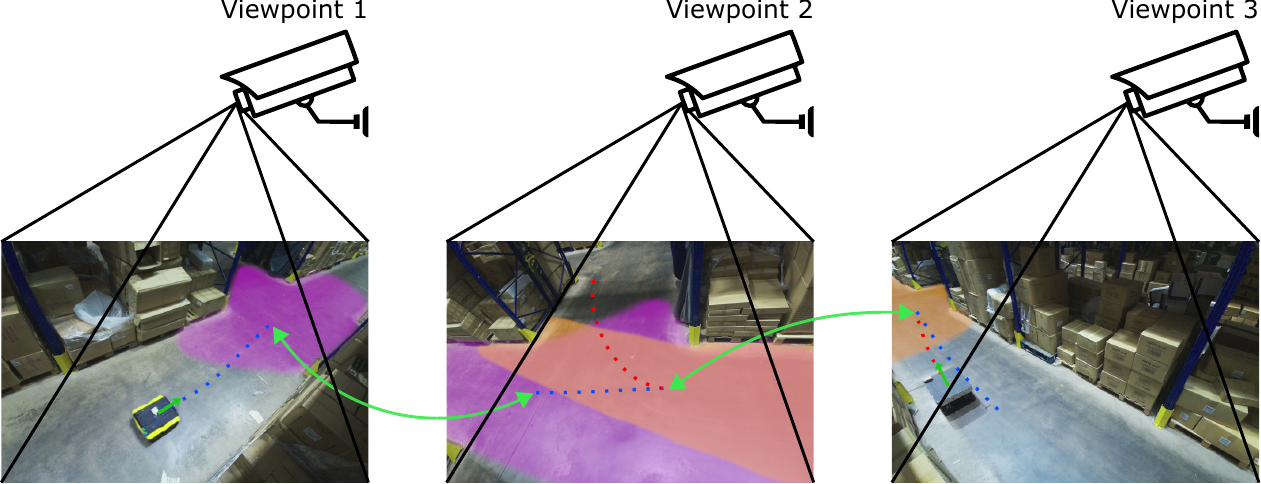}
    \caption{Our setup for deploying multiple robots in a large environment fully covered by CCTV cameras with overlapping viewpoints (coloured regions). Off-boarding all robot compute and planning lets us deploy simple and cheap \emph{blind} robots that only follow received control commands, making deployments more scalable, robust, and practical. Unlike previous works, we support multiple robots through joint planning and control, letting us deploy and validate at larger scale.}
    \label{fig:setting-overview}
\end{figure*}

Real-world deployments, however, must cover large operational areas.
In warehouses this means long, narrow aisles between high shelving that no single vantage point can observe, so practical systems need networks of cameras with overlapping fields of view \cite{batalin2004mobile,poduri2004constrained,robinson2023robot} for robust operation.
Such multi-camera coverage often already exists as part of CCTV installations; where it does not, cameras can be added incrementally to meet the coverage and overlap requirements.
Using these networks for robot control introduces its own challenges, though: many approaches assume careful camera placement, known extrinsics, or global metric alignment~\cite{donmez2019eye,whitaker2020decentralised}, which raise deployment and maintenance costs as cameras are added, replaced, or knocked out of alignment.
These costs grow at warehouse scale, where keeping a large, heavily used network calibrated is a substantial task.

Prior work has shown that a single mobile robot can be navigated through an uncalibrated camera network by representing the environment as a topological camera graph and planning directly in the image plane \cite{robinson2023iros,robinson2025auro}.
This gives reliable off-board navigation without explicit inter-camera calibration, but scaling from one robot to a fleet introduces new challenges. In narrow aisles robots obstruct each other, must pass and yield, and can deadlock in shared regions.
This prior work was also confined to a controlled lab or office, leaving open how robustly the paradigm transfers to industrial settings, whose operational requirements constrain variables such as camera placement.

In this work we extend camera-network off-board control to coordinated multi-robot fleets in a real warehouse.
The system operates entirely in image space over a pixel-wise topological camera graph whose edges are handover regions between overlapping cameras (\cref{fig:setting-overview}).
Within each view, a local planner generates time-parameterised $(x, y, \theta)$ trajectories using a Hybrid A* search with constant-duration motion primitives.
To handle perspective without explicit calibration, we learn per camera-robot models that predict expected image-space motion and robot scale, so the planner reasons consistently about time and space in each view.

Multi-robot coordination uses a hierarchical strategy that escalates only when needed. Robots are first planned independently, treating others as dynamic obstacles, and conflicts are resolved by jointly planning only the robots involved.
Camera overlap regions are treated as shared resources (\emph{semaphores}) owned by at most one robot at a time, preventing collisions and deadlocks at handover.
Crucially, the platforms carry no meaningful onboard sensing or computing for navigation; all perception, planning, and control run on edge compute connected to the network.

\begin{figure}
    \centering
    \begin{subfigure}{0.49\linewidth}
        \centering
        \includegraphics[height=4.2cm]{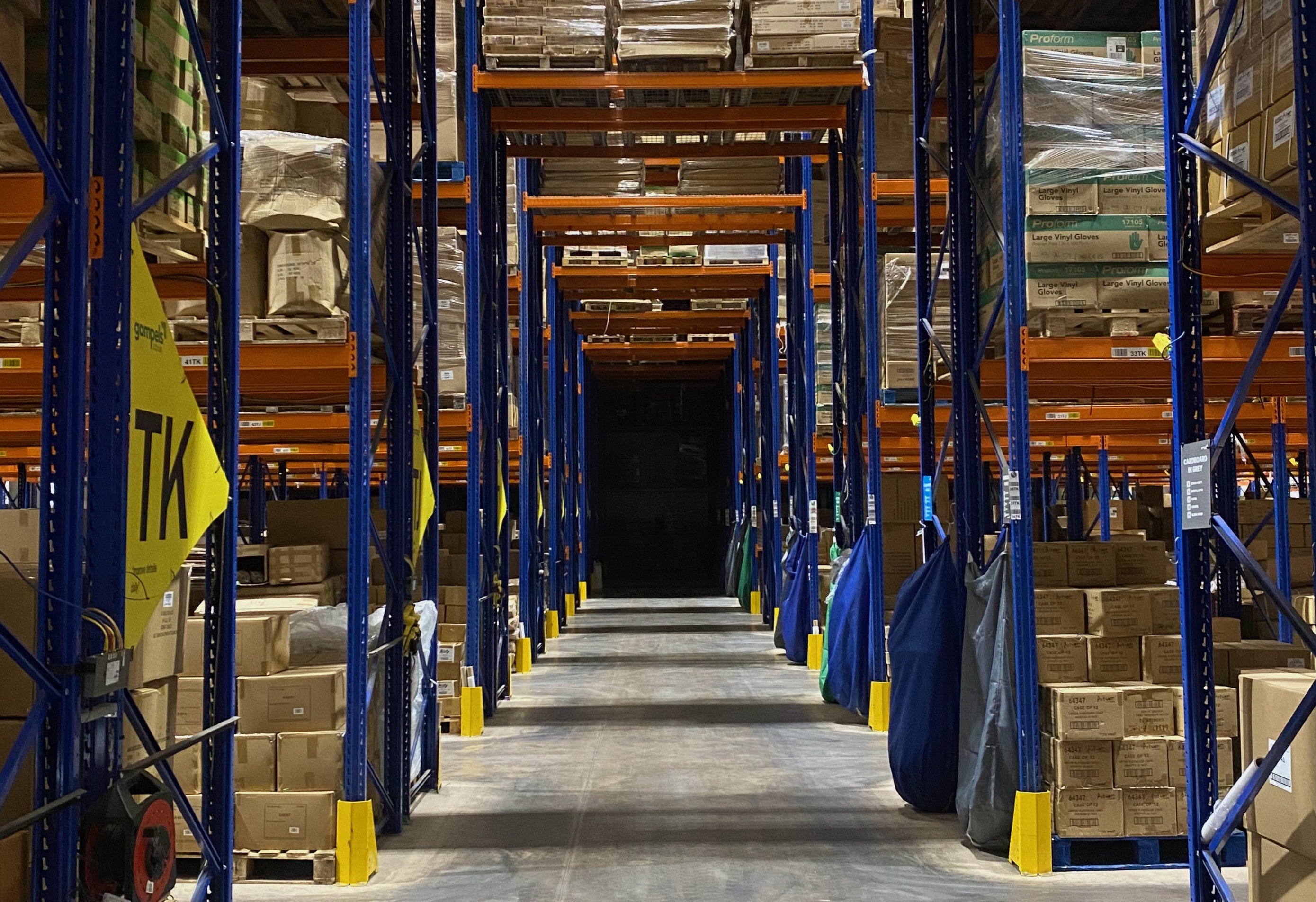}
        \caption{Warehouse deployment environment.}
        \label{fig:warehouse_scale}
    \end{subfigure}
    \hfill
    \begin{subfigure}{0.49\linewidth}
        \centering
        \includegraphics[height=4.2cm]{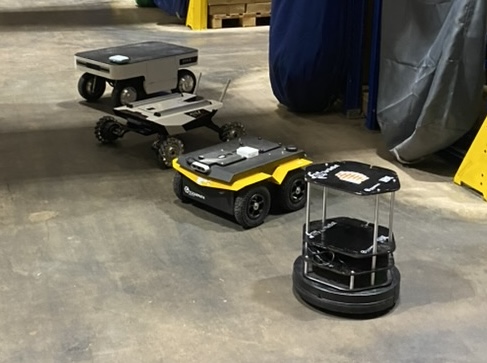}
        \caption{Team of robots that we deploy.}
    \end{subfigure}
    \caption{We validate our system in a large-scale deployment in a warehouse. We rely on 30 CCTV cameras to cover the six 27 m warehouse aisles and deploy simultaneously four different robots.}
    \label{fig:deployment_vis}
\end{figure}

We validate the approach outside the lab in a mid-sized warehouse with four heterogeneous robots and 30 fixed cameras covering six aisles of approximately \qty{27}{\meter}~(\cref{fig:deployment_vis}).
Robots run point-to-point missions along and across aisles, compared against human teleoperation.
We report success rates, mission times, and conflict statistics, and ablate the planning stages.
To our knowledge, these are the first field results for multi-robot planning and coordination in a warehouse using only an external camera network and off-board compute, with robots carrying no task-specific sensing or navigation hardware.

\section{Background and Related Work}
\label{sec:related_work}
\subsection{Infrastructure-Based Visual Servoing for Mobile Robots}

Within robot control, the most relevant area is eye-to-hand visual servoing, where fixed external cameras observe the robot and supply the control feedback~\cite{chaumette_visual_2006}.
Better known for manipulation, it has also been adapted, in a small but persistent literature, to wheeled robots viewed by external cameras.

Early work focused on stable control under non-holonomic constraints, often with overhead or otherwise favourable camera placement.
Dixon et al.~\cite{dixon_adaptive_2001} showed adaptive tracking from an uncalibrated camera, though under alignment assumptions; later work relaxed parameter requirements or improved the controller while still typically assuming strict placement~\cite{qingsong_tracking_2007,yang_adaptive_2018}.
More recent methods use less constrained viewpoints, including purely image-based visual servoing from an uncalibrated external camera, but remain largely single-camera~\cite{liang_purely_2020}.
A key insight from these works is that running the feedback loop in image space rather than robot space removes the need for precise camera calibration while keeping accurate control; this is known as \gls{ibvs}~\cite{dixon_adaptive_2001}.

The multi-camera literature extends this from local control to wide-area operation, but usually adds constraints.
Some works assume cameras aligned with the ground plane so images can be stitched together~\cite{donmez2019eye,whitaker2020decentralised}, while others target world-frame state estimation, often combining infrastructure cameras with onboard sensing or explicit geometric reconstruction~\cite{poornima_robust_2023,shim_mobile_2016}.
Others use infrastructure sensors for global situational awareness in cluttered or crowded spaces rather than for direct robot control~\cite{ravankar2016intelligent,kim_cctv-informed_2024}.
Across these, multi-camera visual servoing of ground robots typically assumes more about calibration, viewpoint geometry, or sensing infrastructure than the single-camera case.

These assumptions become increasingly undesirable at warehouse scale, where cameras are numerous, added incrementally, replaced over time, and prone to drift or disturbance, so maintaining precise extrinsic calibration or global metric alignment adds ongoing deployment and maintenance burden.

Recent work instead carries the calibration-free image-based methods of single-camera servoing over to the multi-camera setting~\cite{robinson2025auro}.
Transitions between camera frames are handled by defining a set of corresponding pixels across images, each known to map to a pixel in another camera, which forms a topological camera graph for moving beyond a single \gls{fov}.
This work builds on that idea, extending it to the multi-robot scenario.

\subsection{Coordination for Multi-Robot Systems}

Once robots share camera views and handover regions, coordination becomes necessary.
The most relevant ideas come from the \gls{mapf} literature, which studies how multiple agents can move from start to goal without colliding in shared space~\cite{stern_multi-agent_2022}.
Classically, \gls{mapf} is posed as a graph search, with methods differing mainly in how explicitly they reason about interactions between agents.

At one extreme, joint planning considers all robots at once. It is optimal but quickly becomes intractable, as the branching factor grows exponentially with the number of robots.
At the other, prioritised planning orders the robots and plans them sequentially, treating higher-priority robots as moving obstacles for the rest~\cite{silver_cooperative_2005}.
This is simple and efficient but incomplete, and can fail in tightly constrained environments.

Between these extremes, some methods expand to joint planning only where conflicts arise: subdimensional expansion raises the search dimensionality only for the robots involved in a local interaction~\cite{wagner_m_2011}, while \gls{cbs} separates individual path generation from higher-level conflict resolution~\cite{sharon_conflict-based_2015}.
Of these, \gls{cbs} is by far the most popular, with many extensions in the literature~\cite{boyarski_icbs_2015,andreychuk_improving_2021}.

Here these methods are relevant as coordination mechanisms rather than as the central research topic.
We do not address general \gls{mapf}; instead we adapt its core ideas inside a multi-camera visual-servoing system whose planning problem is shaped by image-space control, camera handover, and the kinematic constraints of mobile robots.
Coordination thus provides a lightweight mechanism for collision avoidance, conflict resolution, and deadlock prevention, while the main contribution remains calibration-light, infrastructure-based multi-robot control over a camera network.

\section{System Description} \label{sec:system_overview}

We consider a fleet of mobile robots in a large indoor environment, such as a warehouse, monitored by fixed CCTV cameras whose fields of view collectively cover the space.
The robots carry no onboard sensors or computers for navigation: all perception, planning, and control run off-board on edge servers, using only the video streams as feedback.
The goal is to navigate several robots simultaneously between arbitrary start and goal locations in the facility.

Building on the single-robot architecture of \cite{robinson2025auro}, we extend the paradigm to coordinated fleets.
As in our prior work, the stationary, uncalibrated cameras are related through a topological \emph{camera graph} that captures their \gls{fov} overlaps, so perception, velocity estimation, and trajectory planning all run on the 2D image plane of individual cameras, with the graph handling transitions between views.
The multi-robot setting, however, requires robots to share physical space and camera resources without colliding or deadlocking, shifting the problem from purely spatial path planning to spatiotemporal trajectory planning with explicit time and velocity, and prompting major revisions to the hierarchical planner.
The rest of this section covers the one-time setup data-driven models that replace explicit calibration (\cref{sec:deployment_pipeline}) and then the system's sustained operation (\cref{sec:deployment_and_operation}).

\subsection{Preparation Before Deployment}
\label{sec:deployment_pipeline}

Deployment requires a one-time setup to initialise the environment representation and perception models.
Because the system needs no metric calibration, this reduces to five lightweight steps: camera installation, graph construction, driveable-region segmentation, robot perception learning, and scale and velocity learning.
Most are inherited from our prior single-robot system~\cite{robinson2025auro} and summarised only briefly; we focus on the two additions for multi-robot operation: per-edge orientation offsets and image-space velocity prediction.

First, cameras are positioned manually so that their combined fields of view cover the operating area with enough pairwise overlap for handovers; no intrinsic or extrinsic calibration is required, and the mounting angle is unconstrained.

We then build a topological graph whose nodes are cameras and whose edges are overlap regions, each storing a \emph{handover point} (a pixel in each view corresponding to the same physical location) as in \cite{robinson2025auro}.
For multi-robot control we additionally annotate every edge with an approximate, constant in-image orientation offset between the two views at the handover point, allowing us to pass both position and orientation information between the camera views.
We found this sufficient for smooth, controllable transitions without metric calibration.

Following \cite{robinson2025auro}, the driveable floor in each view is segmented once at installation with a prompted foundation segmentation model (SAM2~\cite{ravi2024sam2}), and all remaining pixels are treated as static obstacles.

Per-camera detection and image-space orientation estimation follow \cite{robinson2025auro} unchanged: a per-camera YOLOv8 detector, trained from a small set of labelled in-place spins, returns the robot's bounding box, and a regressor, trained on automatically pseudo-labelled straight-line random walks, returns the heading (see \cite{robinson2023iros} for full details).
We assume each robot is visually distinct and train the detector to tell them apart.

Finally, planning in time as well as space requires the planner to anticipate the robot's apparent size and speed at any image location.
For each robot-camera pair we train a small MLP mapping a pose \((x, y, \sin\theta, \cos\theta)\) to bounding-box dimensions and image-space speed under an MSE loss, reusing the random-walk data; inputs and targets are normalised to \([-1, 1]\) for resolution invariance and rescaled to pixels at inference, so the model implicitly captures each view's perspective and scale with no 3D calibration.
The size branch reuses the prediction model of \cite{robinson2025auro} (previously used only for obstacle expansion), while the speed branch is new and supplies the time parameterisation the multi-robot planner (\cref{sec:low_level_planning}) needs.
Predictions are precomputed over a dense pose grid into a lookup table, so the planner incurs no per-query inference cost.

\subsection{Deployment and Operation}
\label{sec:deployment_and_operation}

\begin{figure*}
    \centering
    \includegraphics[width=0.8\linewidth]{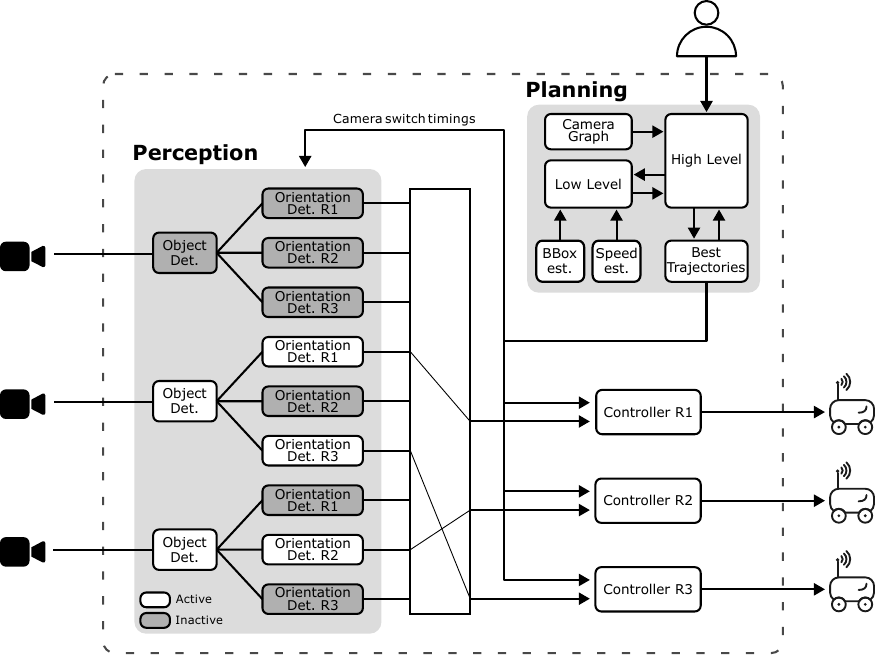}
    \caption{The proposed off-board multi-robot control architecture. Robots with no onboard navigational compute receive control commands over WiFi from an edge server, which processes live video from an uncalibrated CCTV network, performs robot detection and state estimation, and coordinates safe navigation through a hierarchical planner operating on an image-space topological camera graph.}
    \label{fig:system-overview}
\end{figure*}

At mission time, an operator specifies a goal pixel in some camera for each robot.
The system first localises every robot by running the detector and orientation estimator on one frame per camera until all robots are found and their poses estimated.
It then plans and executes trajectories through the hierarchical planner of \cref{fig:system-overview}: a high-level stage selects the sequence of cameras to traverse and keeps an up-to-date record of the low-level trajectories in each; a low-level stage produces, on request, a kinematically feasible image-space trajectory within a camera, free of local collisions with obstacles and other robots; and a per-camera controller drives each robot over Wi-Fi towards its next handover point.

Because the expensive low-level stage decomposes per camera, its planning state space is bounded to a single camera's \gls{fov} and can be parallelised across edge nodes.
As each mission is composed of per-camera legs, cameras can also be activated only when needed to conserve resources. The high-level planner decides when cameras switch on and off and when control is handed over, from each robot's planned position over time (\cref{fig:system-overview}).

\subsubsection{High-Level Planning}
\label{sec:high_level_planning}

The high-level planner is a central process interfacing with both the operator and the low-level planner (\cref{alg:multirobot_planning}).
For each new goal it first selects a sequence of cameras to traverse using Dijkstra's algorithm on the camera graph, as in \cite{robinson2025auro}, where we refer to each camera in this sequence as a \emph{leg}.
It then sequentially calls the low-level planner to find feasible in-image routes for each leg along the high-level path, additionally passing in the trajectories of other robots planned to be in the camera at the same time and therefore providing a possible interaction.
If planning one robot's trajectory in a camera alters another robot's, the high-level planner replans all of the affected robot's future legs, since its handover-point arrival time may have changed.

\begin{algorithm}[t]
    \caption{Hierarchical Multi-Robot Planning on Camera Graph.}
    \label{alg:multirobot_planning}
    \begin{algorithmic}[1]
        \REQUIRE Robot $r$, start camera $c_s$, start pose $\mathbf{p}_s$, start time $t_s$, goal camera $c_g$, goal position $\mathbf{g}$
        \ENSURE Collision-free trajectories for all affected robots

        \STATE $\mathcal{C} \leftarrow$ Shortest path from $c_s$ to $c_g$ by node count
        \STATE $\mathcal{L}_{\text{plan}} \leftarrow \{(r, c_i, \mathbf{g}_i)\}_{i=1}^{|\mathcal{C}|}$ \COMMENT{Legs requiring trajectory planning}
        \STATE $\mathcal{T} \leftarrow \emptyset$ \COMMENT{Completed trajectories}

        \WHILE{$\mathcal{L}_{\text{plan}} \neq \emptyset$}
        \STATE $(r_*, c_*, \mathbf{g}_*) \leftarrow \textsc{SelectNextLeg}(\mathcal{L}_{\text{plan}})$
        \STATE $\mathcal{T}_{c_*} \leftarrow \text{existing trajectories in camera } c_*$

        \STATE $\mathcal{R}_{\text{joint}} \leftarrow \{r_*\}$ \COMMENT{Start with prioritised planning}
        \REPEAT
        \STATE $\mathcal{T}_{\text{new}} \leftarrow \textsc{JointPlanning}(\mathcal{R}_{\text{joint}}, \mathcal{T}_{c_*}, c_*)$
        \IF{planning failed}
        \STATE Add one conflicting robot from $\mathcal{T}_{c_*}$ to $\mathcal{R}_{\text{joint}}$
        \ENDIF
        \UNTIL{planning succeeded or planner runs out of time}

        \FOR{each replanned robot $r_i \in \mathcal{R}_{\text{joint}} \setminus \{r_*\}$}
        \STATE Add downstream legs of $r_i$ to $\mathcal{L}_{\text{plan}}$ \COMMENT{Must replan future cameras}
        \ENDFOR

        \STATE Update $\mathcal{T}$ with $\mathcal{T}_{\text{new}}$ and remove completed legs from $\mathcal{L}_{\text{plan}}$
        \ENDWHILE

        \RETURN $\mathcal{T}$
    \end{algorithmic}
\end{algorithm}

\subsubsection{Low-Level Planning}
\label{sec:low_level_planning}

Within each leg, we plan in image space from the entry handover point (or the robot's current position) to the exit handover point (or final goal).
The initial orientation is derived from the robot's pose at the previous exit handover point and the per-edge orientation offsets (\cref{sec:deployment_pipeline}).
The planner is the Hybrid A* search of \cite{robinson2025auro} over poses \((x, y, \theta)\) within the driveable region, reformulated for multi-robot use. Where \cite{robinson2025auro} advances each motion primitive a fixed \emph{distance} in pixels, we now advance it a fixed \emph{duration}, setting its pixel length per step from the predicted image-space speed (\cref{sec:deployment_pipeline}) so that constant ground speed is maintained despite perspective.
Every search node therefore corresponds to a known time, which enables collision detection between robots.
We use six primitives: three forward arcs, two turn-on-the-spot rotations, and a `wait'.
Per-pose collision checking (the predicted bounding box, expanded by a constant factor, checked against the driveable region) and the obstacle-proximity cost term are retained from \cite{robinson2025auro}, and the Hybrid A* cost penalises distance travelled and waiting or non-forward actions to encourage progress towards the goal.

\subsubsection{Multi-Robot Coordination}
\label{sec:multi-robot_coordination}

Robots are planned one at a time in a \emph{prioritised-then-joint} scheme.
A new robot is first planned with all already-planned robots treated as moving obstacles occupying their predicted bounding boxes along their trajectories, which resolves most interactions cheaply.
Where it cannot, typically in a narrow aisle where both robots must give way, the planner escalates and replans both jointly (\cref{alg:multirobot_planning}).
As in subdimensional expansion~\cite{wagner_m_2011}, this bounds the expensive joint planning to the leg on which the conflict occurs.

Camera overlap regions (a fixed region around the handover point) need special care, since a robot crossing one is briefly visible to, and controllable by, two cameras.
We treat each as a semaphore held by at most one robot at a time: while it is occupied, it is marked as an obstacle for all other robots in both cameras, which wait at its boundary until it clears.
This prevents the collisions that could otherwise occur when robots controlled by separate cameras share that space.

\subsubsection{Control}
\label{sec:control}

Trajectories are followed in image space by a controller on the edge compute that sends command velocities over Wi-Fi.
We use a pure-pursuit controller~\cite{coulter1992implementation} for forward motion, with the pursuit point advancing along the planned trajectory, and a PD controller to turn the robot on the spot for rotation primitives.

\section{Experiment Setup}
\label{sec:experimental_setup}

We validated the full system in a real, working warehouse with four heterogeneous robots and a 30-camera network.
The deployment both recorded data for the component-level study (\cref{sec:val_pred_models}) and evaluated coordinated operation end-to-end (\cref{sec:val_full_system}).

\subsection{Environment and Camera Network}

\begin{figure}[h]
    \centering
    \includegraphics[width=0.75\linewidth]{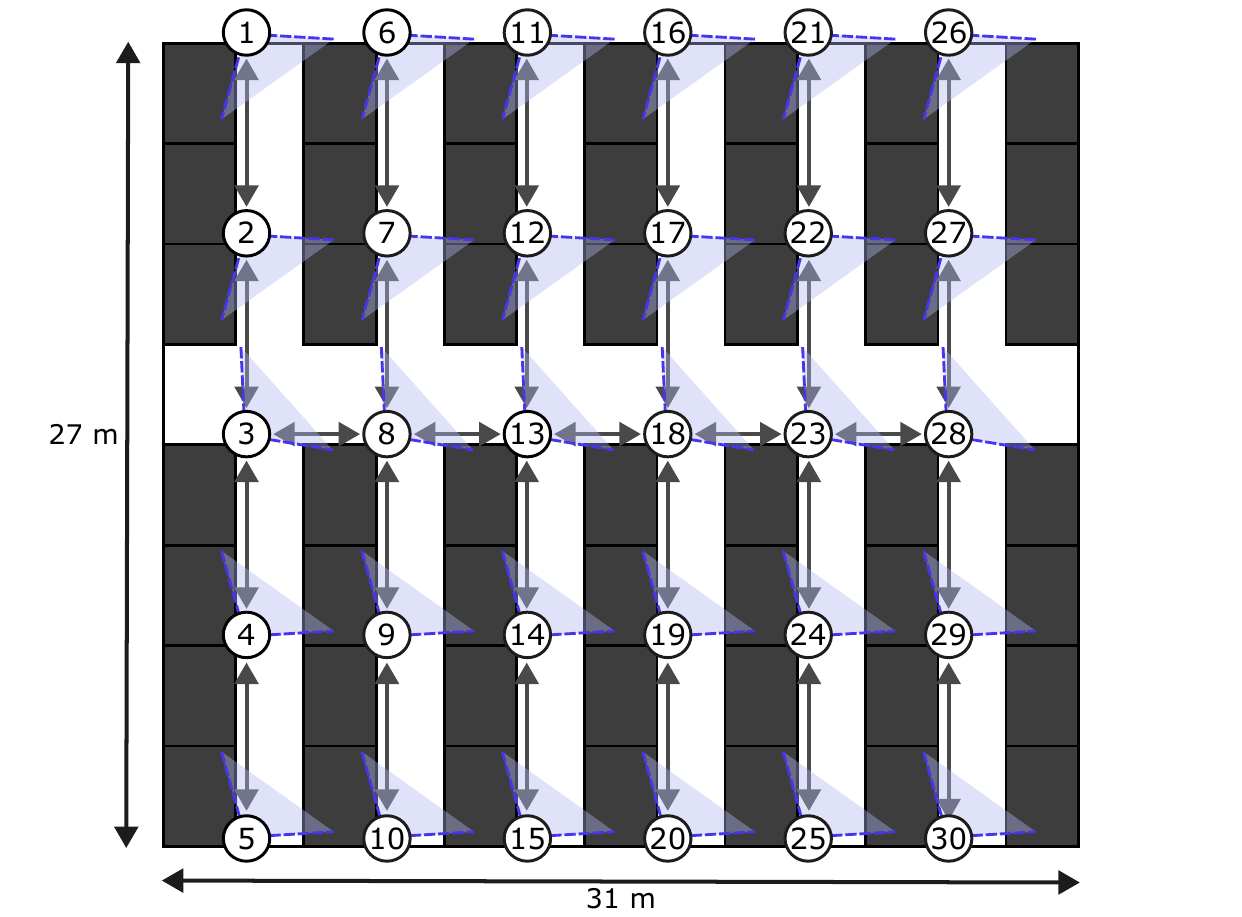}
    \caption{Layout of the mid-sized warehouse used for evaluation: six parallel aisles monitored by 30 overhead RGB cameras that fully cover the space with overlapping fields of view (with this relationship here shown by arrows). Because the aisles are symmetric, the data-driven perception models and graph topology trained on one primary aisle transfer directly to the structurally identical regions elsewhere.}
    \label{fig:warehouse_plan}
\end{figure}

Experiments were run in a mid-sized warehouse (\cref{fig:warehouse_scale}) of six parallel aisles, each roughly \SI{27}{\meter} long and flanked by shelving, with only robots present in the test area.
We instrumented it with 30 custom cameras (Raspberry Pi 4 Model B with Camera Module 3, powered over PoE), each streaming at \SI{5}{fps} at either low (\(640\times 420\)) or high (\(1920\times 1260\)) resolution depending on its typical distance to the robots.
Camera placement (\cref{fig:warehouse_plan}) followed practical mounting constraints rather than any robotics-specific optimisation, yet we aimed to reduce the amount of overlap between camera \glspl{fov} to only that needed for camera handover.

We exploited the warehouse's highly structured layout by placing cameras identically in every aisle (aligned roughly by eye).
This confined the costly setup steps (recording perception training data and marking the camera-graph overlap points (\cref{sec:deployment_pipeline})) to a single aisle, after which the trained models and overlap points were reused throughout; every per-camera quantity below was thus recorded only five times (cameras 1--5) rather than thirty.
The driveable region was obtained from a prompted SAM2~\cite{ravi2024sam2} in all thirty cameras, and the camera graph was built by selecting one aisle's five handover points and copying them across the space.
Although \cite{robinson2025auro} gives an automated graph-building method that could be extended with the angle offset, the warehouse's regular structure made it easier to mark edges in one aisle and copy them across.

\subsection{Robots and Computing}

The four platforms, a \emph{Clearpath Turtlebot 2} and \emph{Jackal} and an \emph{AgileX Scout Mini} and \emph{Ranger Mini 3.0}, were chosen to be visually distinct, a constraint we revisit in \cref{sec:limitations}.
Each had its sensors removed and was driven solely by off-board linear- and angular-velocity commands over Wi-Fi.
All perception, planning, and control ran on a single edge server (AMD Threadripper 7960X, NVIDIA RTX 4090); we centralise it here for convenience, but it could equally be distributed across edge nodes.

\subsection{Perception and Prediction Models}

Training data for all learned models was collected as in \cite{robinson2023iros}.
The per-camera YOLOv8-small detector~\cite{yolov8} was trained from manually labelled in-place spins at 30 locations along the training aisle; this large number of spins is needed only to tell the robots apart and could be reduced with a tracking method to distinguish robots (\cref{sec:limitations}).
The orientation regressor and the bounding-box and speed MLPs (\cref{sec:deployment_pipeline}) were trained on \SI{40}{\minute} of automatically pseudo-labelled random walks per robot-camera pair, with speed labels taken from successive bounding-box centres, smoothed with a three-frame moving average, and with the acceleration phases at each straight line's start and end removed.
The two prediction MLPs share an architecture of three hidden layers \([128, 64, 32]\) with ReLU activations, both selected by cross-validation.
YOLOv8s used the Ultralytics defaults; the orientation regressor was trained for 20 epochs at learning rate \(10^{-4}\) and batch size 64, with a 90/10 train/validation split; the two prediction MLPs used dropout \(0.1\), 400 epochs, learning rate \(10^{-3}\), and batch size 32.

\subsection{Evaluation Scenarios}

With the system calibrated and its controller gains and planning heuristics fixed, we ran three experiments.
First, a component-level study validated the box-size and speed predictors across the random-walk data (\cref{sec:val_pred_models}).
Second, an extended-autonomy test, in which all four robots navigated continuously to randomly drawn goals across the whole space, exposed robustness and failure modes (\cref{sec:extended_full_autonomy}).
Finally, a teleoperation comparison measured performance against a human operator: one robot shuttled between two fixed points (autonomously and under teleoperation), alone and then with two others running random routes as dynamic obstacles (\cref{sec:teleop}).

\section{Results}
\label{sec:results}

In this section, we first evaluate the accuracy of the predictive models for bounding box sizes and velocities (\cref{sec:val_pred_models}), before detailing our findings from deploying and coordinating the full multi-robot system (\cref{sec:val_full_system}) and comparing it against human teleoperation (\cref{sec:teleop}).

\subsection{Validating the Bounding Box Size and Velocity Prediction Models}
\label{sec:val_pred_models}

We validate our bounding-box-size and speed predictors (\cref{sec:deployment_pipeline}) on the recorded random walks, whose pseudo-labels of position and velocity come from the robot detector.
We use 3-fold cross-validation over legs to evaluate on unseen trajectories, while the deployed models are retrained on all legs for maximum coverage.
We compare against two baselines.
The \emph{Mean} baseline always predicts the training-set average, while the \emph{Nearest-Neighbour} baseline returns the value of the training sample with the closest pose $(x, y, \theta)$.
This isolates the benefit of our learning-based smoothing over direct retrieval from noisy data.
Thanks to the limited size of our environment, only a few minutes of recorded legs per pair were enough to outperform both.

\begin{figure*}[t]
    \centering
    \includegraphics[width=\linewidth]{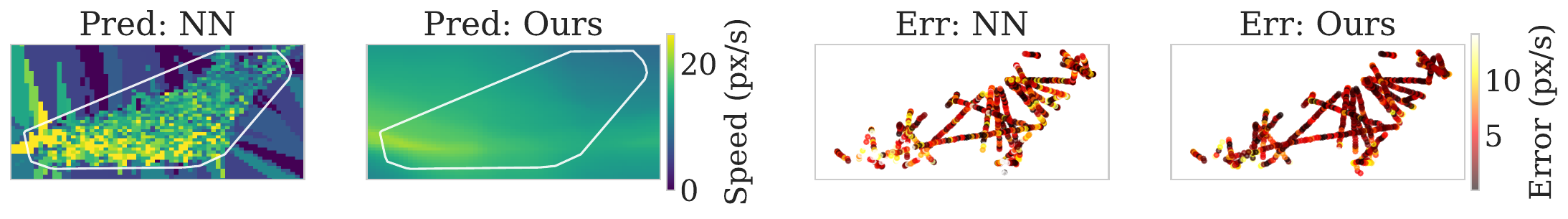}
    \caption{Image-space speed for the nearest-neighbour baseline and our method: dense predictions across the image plane (left two panels; fixed orientation $\theta = 0\degree$, white outline marks the training-data boundary) and per-trajectory error on held-out data (right two panels). Our method is smooth and perspective-consistent with low, uniform error, whereas the baseline is discontinuous where training data is sparse.}
    \label{fig:speed_predictions}
\end{figure*}

\Cref{fig:speed_predictions} compares image-space speed predictions for the two methods.
The nearest-neighbour baseline is discontinuous and interpolates poorly where data is sparse.
Our method is smooth and perspective-consistent, with image-plane speed decreasing as the robot moves further away from the camera, and keeps consistently low error across the unevenly covered driveable area.

\begin{figure*}[t]
    \centering
    \includegraphics[width=1.0\linewidth]{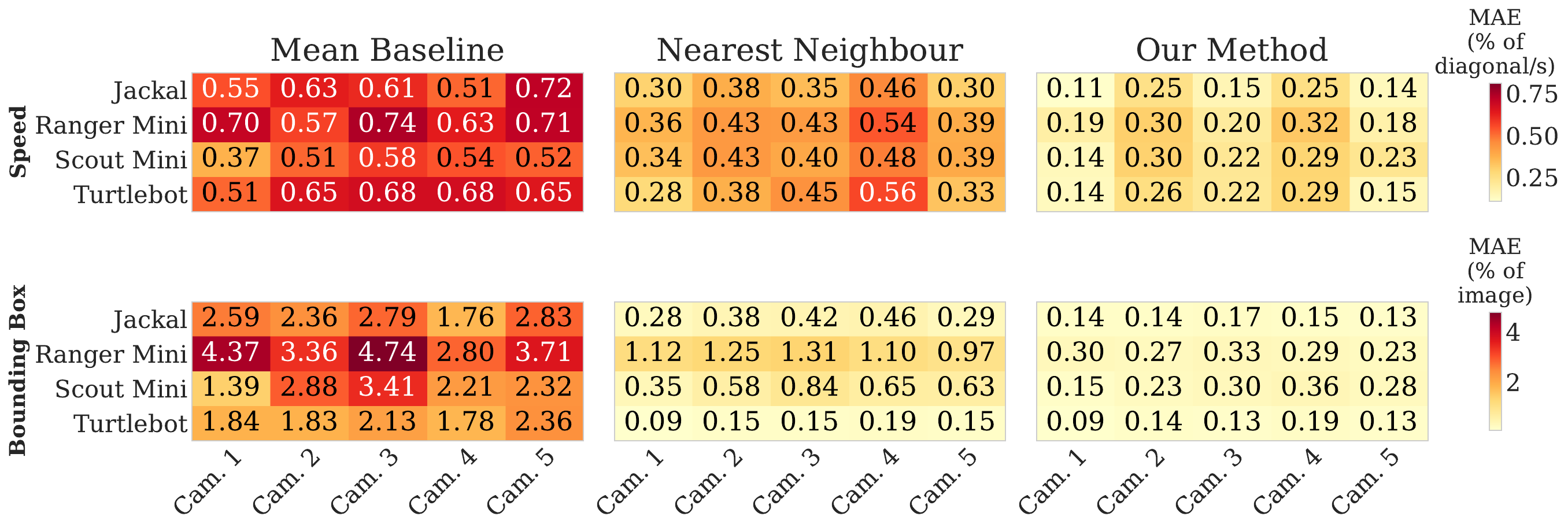}
    \caption{Mean absolute error (MAE), relative to image dimensions, for bounding-box and speed predictions across all cameras and robots.
    }
    \label{fig:error_heatmaps}
\end{figure*}

\Cref{fig:error_heatmaps} quantifies this across all robot and camera pairs, reporting mean absolute error (relative to image dimensions) for both box sizes and speeds.
Our method stays consistently low and outperforms both baselines.
Errors peak at the most challenging cameras, such as 2 and 4 with their extreme perspectives or occlusions, and are lower for the smaller, more distinctive Jackal and Turtlebot.
Errors generally remain below 0.5\% of the corresponding image dimension, which is within 10 pixels at $1920 \times 1080$ resolution and sufficient for reliable planning.

\subsection{Validating the Full System: Extended Full Autonomy}
\label{sec:val_full_system}
\label{sec:extended_full_autonomy}

\begin{figure}
    \centering
    \begin{subfigure}{0.49\linewidth}
        \centering
        \includegraphics[width=\linewidth]{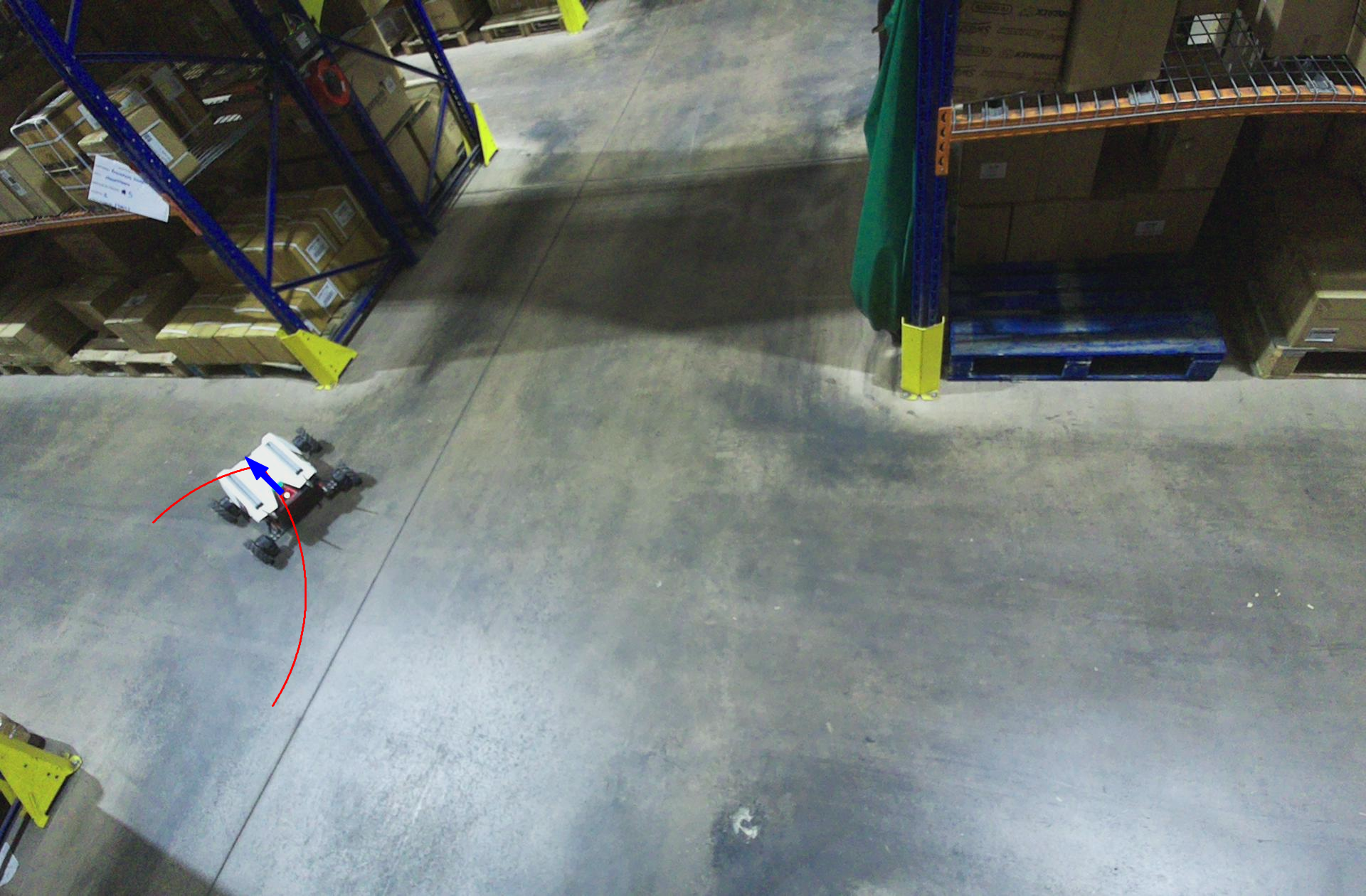}
        \caption{Camera 8}
    \end{subfigure}
    \begin{subfigure}{0.49\linewidth}
        \centering
        \includegraphics[width=\linewidth]{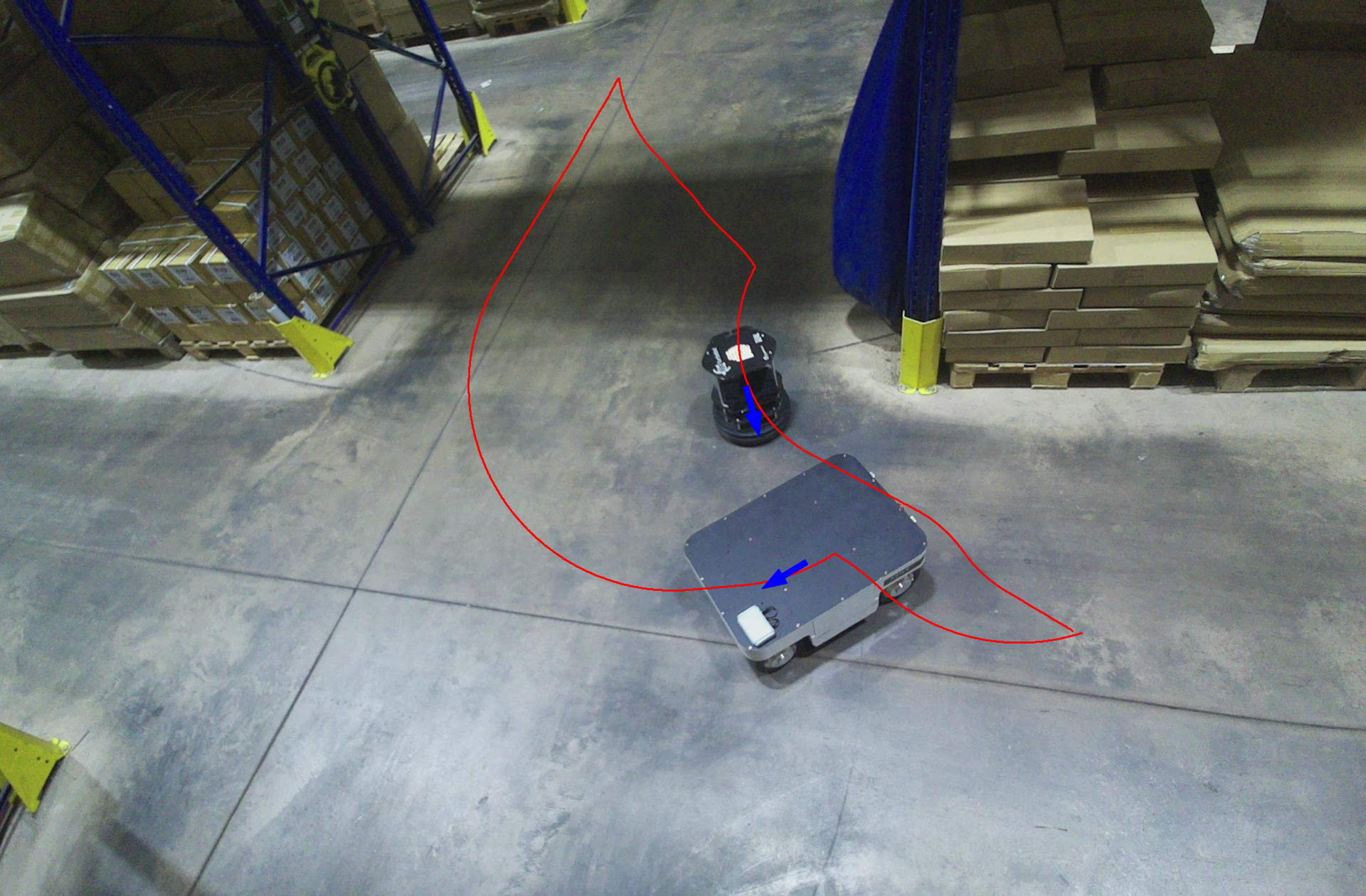}
        \caption{Camera 18}
    \end{subfigure}
    \caption{View of three of the four robots at one instant of the end-to-end validation test, where each robot is shown from the perspective of the camera controlling it. Red lines indicate the path planned for each robot, and blue arrows show the detected in-image pose.}
    \label{fig:whole_system_example}
\end{figure}

As an end-to-end validation, four robots navigated the test environment simultaneously. Each was assigned a random goal position among the 30 camera views and, on reaching it, was immediately given another, giving continuous multi-target operation (\cref{fig:whole_system_example}). The system ran for \SI{45}{\minute}, over which the fleet traversed \SI{1.3}{\kilo\meter}.

The fleet completed 76 point-to-point missions. On reaching a goal, each robot idled for two seconds to conservatively cover the empirical worst-case planning time for its next route, yet the cumulative planning time over the whole test was only \SI{1.06}{\second} ($\sim\SI{14}{\milli\second}$ per plan). With the robots sparsely distributed across the large workspace, almost all conflicts were resolved by the faster prioritised planning, without falling back to joint planning.
The fixed two-second buffer therefore caused unnecessary idle time. This is only a minor inefficiency on long routes, but it is a limitation of the current implementation that could be removed by planning the next mission before the current one ends.

Infrastructure use was also efficient: despite covering a broad area, on average only 3.5 cameras were streaming and computing at any moment. This shows how off-board, shared-resource control avoids the redundant per-robot resource use typical of onboard autonomy.

As a prototype built under tight warehouse-access constraints, the system needed four interventions, at minutes 10, 17, 37, and 41, all of which would be easily avoided in production. The first two were collisions after a robot failed to track a maximum-curvature segment; tuning its controller gains on the fly after the second (at \SI{17}{\minute}) resolved this for the rest of the deployment.
The last two were perception-driven: at \SI{37}{\minute} the detector transiently misclassified a robot, and at \SI{41}{\minute} a robot crossed a visual region underrepresented in training, giving an erratic orientation estimate; both triggered a pose-update timeout that halted the controller. As also seen in \cite{robinson2023iros}, data-driven pose estimation degrades where training data is sparse, so a real deployment must ensure the whole operating area is well covered. Random walks are not the only way to achieve this: data can be gathered continually while robots perform useful tasks, for example under teleoperation.

\subsection{Comparison to Human Teleoperation}
\label{sec:teleop}

In a second experiment we compared the efficiency of autonomous control against human teleoperation. The Jackal was driven from a fixed point in camera 2 to a fixed point in camera 4 and back, both by our system and by a human operator using the camera feeds. Each case was run either with the Jackal alone or with two other robots roaming the area on random point-to-point paths, which could cross the Jackal's route and so test coordination.

Travel times are reported in \cref{tab:teleoperation_results}; the autonomous durations include the \SI{2}{\second} stationary planning period before each of the two legs. The Jackal relied almost entirely on prioritised planning, never exceeding \SI{20}{\milli\second}, so this \SI{4}{\second} of conservative waiting inflates the reported times; a less conservative setting would remove most of it and bring the autonomous times close to the teleoperation baseline.

Joint planning was triggered only once, when a background robot could not bypass the Jackal's planned route, in the slowest two-robot autonomous run of \SI{124}{\second}. The Jackal was re-planned jointly en route without stopping, as part of the background robot's planning, causing it to divert its track.
This took \SI{1513}{\milli\second} but ran while the Jackal was still in the previous camera, so it did not affect its time.


\begin{table}
    \centering
    \setlength{\tabcolsep}{6pt}
    \caption{Time taken (seconds) for the Jackal to travel from camera 2 to camera 4 and back, with different numbers of additional robots moving in the environment. For each method, we report individual trial results with their mean and standard deviation.}
    \begin{tabular}{ccccc}
        \toprule
        \multirow{2}{*}{\shortstack{Additional\\robots}} & \multicolumn{2}{c}{Autonomous} & \multicolumn{2}{c}{Teleoperation} \\
        \cmidrule(lr){2-3} \cmidrule(lr){4-5}
         & Trials & Mean $\pm$ std & Trials & Mean $\pm$ std \\
        \midrule
        0 & 110.5, 110.6, 111.4 & $110.8 \pm 0.4$ & 109.2 & $109.2 \pm 0.0$ \\
        2 & 116.2, 124.0, 113.0 & $117.7 \pm 4.6$ & 110.8, 117.1, 111.1 & $113.0 \pm 2.9$ \\
        \bottomrule
    \end{tabular}
    \label{tab:teleoperation_results}
\end{table}


With no other robots, the autonomous system was on average 5.1\% slower than teleoperation including planning time, and only 1.5\% slower without it. Adding two robots made the times much more variable, with a significantly larger standard deviation: sometimes the other robots left a clear path, and sometimes the Jackal had to deviate. The human stayed close to the single-robot baseline more consistently.

Overall, the autonomous system stayed competitive with a human operator while also controlling several robots in parallel across different camera views, which we found too demanding for a single operator to manage.

\section{Limitations and Future Work}
\label{sec:limitations}

Despite our promising results, several limitations remain in our current approach.
The most immediate is that robots stay stationary while planning, because the planner needs a fixed start time before it can begin.
Much of this idle time could be recovered by planning each robot's next mission before it finishes the current one, but still \cref{sec:extended_full_autonomy} showed that the full worst-case budget is rarely needed, so other solutions will need to be explored.
The controllers were also a source of fragility, needing substantial tuning to run reliably and causing failures when mistuned; more robust control would address this without affecting the rest of the approach.
Two modelling assumptions limit generality: that robots are visually distinct, and that they move at constant ground speed.
The first of these could be relaxed by tracking robots across frames rather than relying on appearance alone, and the second by a planner and controller that actively manage velocity.
A final limitation is geometric: because any overlap between a robot's bounding box and an obstacle counts as a collision, the check is overly conservative for cameras viewing the floor at a shallow angle, where a tall robot's box can clip an obstacle behind it before it actually collides.
High camera mounts largely avoid this, and it caused no noticeable issues in our experiments, but a fuller solution might involve recent ML models which can infer 3D information from 2D images~\cite{zhong_nerfoot_2024}.

\section{Conclusion}
\label{sec:conclusion}

We have presented a field-deployed system that controls a fleet of robots using only an uncalibrated CCTV network and off-board compute.
Operating purely in image space over a pixel-wise topological camera graph, it localises robots with markerless learned models, predicts their image-space size and speed, and coordinates them through a hierarchical prioritised-then-joint planner with semaphore-guarded handovers.
Deployed in a real warehouse with 30 cameras and four heterogeneous robots, the system sustained extended autonomous operation and came within 5.1\% of a human teleoperator on a like-for-like route while controlling several robots in parallel, which a single operator could not manage.
Addressing the limitations above, particularly anticipatory planning, should close this gap further.

\section*{Acknowledgment}
We would like to thank the engineers Pratik Somaiya and Matthew Towlson and our trial manager Daniel Marques for their support during the trials.
We would also like to thank Gompels HealthCare Ltd. for kindly allowing us to use their warehouse facilities.
We adittionaly thank Kyuhwan Yeon and Halil Kelebek for the help they provided in the field.

\bibliographystyle{spmpsci}
\bibliography{biblio}

@inproceedings{donmez2019eye,
    title = {The eye-out-device multi-camera expansion for mobile robot control},
    author = {D{\"o}nmez, Emrah and Kocamaz, Adnan Fatih},
    booktitle = {{IDAP}},
    pages = {1--6},
    year = {2019},
    organization = {IEEE},
}

@article{robinson2025auro,
    author = {Robinson, Luke
              and Gadd, Matthew
              and Newman, Paul
              and Martini, Daniele De},
    title = {Robot-relay: building-wide, calibration-less visual servoing with learned sensor handover networks},
    journal = {Auton. Robots},
    year = {2025},
    month = {Nov},
    day = {28},
    volume = {50},
    number = {1},
    pages = {3},
    issn = {1573-7527},
}

@misc{ravi2024sam2,
    title = {SAM 2: Segment Anything in Images and Videos},
    author = {Nikhila Ravi and Valentin Gabeur and Yuan-Ting Hu and Ronghang Hu and Chaitanya Ryali and Tengyu Ma and Haitham Khedr and Roman Rädle and Chloe Rolland and Laura Gustafson and Eric Mintun and Junting Pan and Kalyan Vasudev Alwala and Nicolas Carion and Chao-Yuan Wu and Ross Girshick and Piotr Dollár and Christoph Feichtenhofer},
    year = {2024},
    eprint = {2408.00714},
    archiveprefix = {arXiv},
    primaryclass = {cs.CV},
}

@inproceedings{robinson2023robot,
    title = {Robot-Relay: Building-Wide, Calibration-Less Visual Servoing with Learned Sensor Handover Networks},
    author = {Robinson, Luke and Gadd, Matthew and Newman, Paul and Martini, Daniele De},
    booktitle = {{ISER}},
    pages = {129--140},
    year = {2023},
    organization = {Springer},
}

@inproceedings{poduri2004constrained,
    title = {Constrained coverage for mobile sensor networks},
    author = {Poduri, Sameera and Sukhatme, Gaurav S},
    booktitle = {{ICRA}},
    volume = {1},
    pages = {165--171},
    year = {2004},
    organization = {IEEE},
}

@inproceedings{batalin2004mobile,
    title = {Mobile robot navigation using a sensor network},
    author = {Batalin, Maxim A and Sukhatme, Gaurav S and Hattig, Myron},
    booktitle = {{ICRA}},
    volume = {1},
    pages = {636--641},
    year = {2004},
    organization = {IEEE},
}

@article{buoso2024select2plan,
    title = {Select2plan: Training-free icl-based planning through vqa and memory retrieval},
    author = {Buoso, Davide and Robinson, Luke and Averta, Giuseppe and Torr, Philip and Franzmeyer, Tim and De Martini, Daniele},
    journal = {{IEEE} Robot. Autom. Lett.},
    volume = {10},
    number = {11},
    pages = {11267--11274},
    year = {2025},
    publisher = {IEEE},
}

@inproceedings{yolov8,
    author = {Varghese, Rejin and Sambath, M.},
    booktitle = {{ADICS}},
    title = {YOLOv8: A Novel Object Detection Algorithm with Enhanced Performance and Robustness},
    year = {2024},
    volume = {},
    number = {},
    pages = {1-6},
}

@article{simoens2018internet,
    title = {The Internet of Robotic Things: A review of concept, added value and applications},
    author = {Simoens, Pieter and Dragone, Mauro and Saffiotti, Alessandro},
    journal = {Int. J. Adv. Robot. Syst.},
    volume = {15},
    number = {1},
    pages = {1729881418759424},
    year = {2018},
    publisher = {Sage Publications Sage UK: London, England},
}

@inproceedings{robinson2023iros,
    title = {{Visual Servoing on Wheels: Robust Robot Orientation Estimation in Remote Viewpoint Control}},
    author = {Robinson, Luke and De Martini, Daniele and Gadd, Matthew and Newman, Paul},
    booktitle = {{IROS}},
    year = {2023},
}

@article{whitaker2020decentralised,
    title = {Decentralised indoor smart camera mapping and hierarchical navigation for autonomous ground vehicles},
    author = {Whitaker, Taylor JL and Cunningham, Samantha-Jo and Bobda, Christophe},
    journal = {{IET} Comput. Vis.},
    volume = {14},
    number = {7},
    pages = {462--470},
    year = {2020},
}

@inproceedings{ravankar2016intelligent,
    title = {Intelligent robot guidance in fixed external camera network for navigation in crowded and narrow passages},
    author = {Ravankar, Abhijeet and Ravankar, Ankit and Kobayashi, Yukinori and Emaru, Takanori},
    booktitle = {ECSA},
    volume = {1},
    pages = {37},
    year = {2016},
    organization = {MDPI},
}

@inproceedings{wagner_m_2011,
    address = {San Francisco, CA},
    title = {M*: {A} complete multirobot path planning algorithm with performance bounds},
    isbn = {978-1-61284-456-5 978-1-61284-454-1 978-1-61284-455-8},
    shorttitle = {M*},
    language = {en},
    booktitle = {{IROS}},
    publisher = {IEEE},
    author = {Wagner, Glenn and Choset, Howie},
    month = sep,
    year = {2011},
    pages = {3260--3267},
}

@article{sharon_conflict-based_2015,
    title = {Conflict-based search for optimal multi-agent pathfinding},
    volume = {219},
    issn = {0004-3702},
    journal = {Artif. Intell.},
    author = {Sharon, Guni and Stern, Roni and Felner, Ariel and Sturtevant, Nathan R.},
    month = feb,
    year = {2015},
    pages = {40--66},
}

@techreport{coulter1992implementation,
    author = {R. Craig Coulter},
    title = {Implementation of the Pure Pursuit Path Tracking Algorithm},
    year = {1992},
    month = {January},
    institution = {Carnegie Mellon University},
    address = {Pittsburgh, PA},
    number = {CMU-RI-TR-92-01},
}

@article{chaumette_visual_2006,
    title = {Visual servo control. I. Basic approaches},
    volume = {13},
    issn = {1558-223X},
    pages = {82--90},
    number = {4},
    journal = {{IEEE} Robot. Autom. Mag.},
    author = {Chaumette, Francois and Hutchinson, Seth},
    year = {2006},
}

@article{dixon_adaptive_2001,
    title = {Adaptive tracking control of a wheeled mobile robot via an uncalibrated camera system},
    volume = {31},
    issn = {1941-0492},
    pages = {341--352},
    number = {3},
    journal = {{IEEE} Trans. Syst. Man Cybern. B Cybern.},
    author = {Dixon, W.E. and Dawson, D.M. and Zergeroglu, E. and Behal, A.},
    year = {2001},
}

@inproceedings{qingsong_tracking_2007,
    title = {Tracking of Nonholonomic Control Systems Based on Visual Servoing Feedback},
    issn = {1934-1768},
    eventtitle = {2007 Chinese Control Conference},
    pages = {459--463},
    booktitle = {{CCC}},
    author = {Qingsong, Li and Chaoli, Wang and Wenbin, Niu},
    year = {2007},
}

@article{yang_adaptive_2018,
    title = {Adaptive and sliding mode tracking control for wheeled mobile robots with unknown visual parameters},
    journal = {Trans. Inst. Meas. Control},
    volume = {40},
    number = {1},
    pages = {269--278},
    author = {Yang, Fang and Su, Hongye and Wang, Chaoli and Li, Zhenxing},
    year = {2018},
    langid = {english},
}

@article{liang_purely_2020,
    title = {Purely Image-Based Pose Stabilization of Nonholonomic Mobile Robots With a Truly Uncalibrated Overhead Camera},
    volume = {36},
    issn = {1941-0468},
    pages = {724--742},
    number = {3},
    journal = {{IEEE} Trans. Robot.},
    author = {Liang, Xinwu and Wang, Hesheng and Liu, Yun-Hui and Liu, Zhe and You, Bing and Jing, Zhongliang and Chen, Weidong},
    year = {2020},
}

@inproceedings{poornima_robust_2023,
    address = {New York, {NY}, {USA}},
    title = {Robust and Scalable Indoor Robot Localization Based on Fusion of Infrastructure Camera Feeds and On-Board Sensors},
    isbn = {978-1-4503-9980-7},
    eventtitle = {{ACM} International Conference Proceeding Series},
    pages = {1--7},
    booktitle = {{AIR}},
    publisher = {Association for Computing Machinery},
    author = {Poornima, J.D. and Krishnapuram, R. and Bharatheesha, M. and Amrutur, B. and Sundaram, S.},
    year = {2023},
}

@article{shim_mobile_2016,
    title = {A Mobile Robot Localization via Indoor Fixed Remote Surveillance Cameras},
    volume = {16},
    issn = {1424-8220},
    pages = {195},
    number = {2},
    journal = {Sensors},
    shortjournal = {Sensors},
    author = {Shim, Jae and Cho, Young},
    year = {2016},
    langid = {english},
}

@article{kim_cctv-informed_2024,
    title = {{CCTV}-Informed Human-Aware Robot Navigation in Crowded Indoor Environments},
    volume = {9},
    issn = {2377-3766},
    pages = {5767--5774},
    number = {6},
    journal = {{IEEE} Robot. Autom. Lett.},
    author = {Kim, Mincheul and Kwon, Youngsun and Lee, Sebin and Yoon, Sung-eui},
    year = {2024},
}

@inbook{stern_multi-agent_2022,
    author = {Stern, Roni},
    title = {Multi-Agent Path Finding – An Overview},
    year = {2022},
    isbn = {978-3-030-33273-0},
    publisher = {Springer-Verlag},
    address = {Berlin, Heidelberg},
    booktitle = {Artif. Intell.},
    pages = {96–115},
    numpages = {20},
}

@article{silver_cooperative_2005,
    title = {Cooperative Pathfinding},
    volume = {1},
    rights = {Copyright (c) 2005 Proceedings of the {AAAI} Conference on Artificial Intelligence and Interactive Digital Entertainment},
    issn = {2334-0924},
    pages = {117--122},
    number = {1},
    journal = {{AIIDE}},
    author = {Silver, David},
    year = {2005},
    langid = {english},
}

@misc{zhong_nerfoot_2024,
    title = {{NeRFoot}: Robot-Footprint Estimation for Image-Based Visual Servoing},
    shorttitle = {{NeRFoot}},
    number = {{arXiv}:2408.01251},
    publisher = {{arXiv}},
    author = {Zhong, Daoxin and Robinson, Luke and Martini, Daniele De},
    year = {2024},
    eprinttype = {arxiv},
    eprint = {2408.01251 [cs]},
}

@inproceedings{boyarski_icbs_2015,
    address = {Buenos Aires, Argentina},
    title = {{ICBS}: improved conflict-based search algorithm for multi-agent pathfinding},
    isbn = {978-1-57735-738-4},
    shorttitle = {{ICBS}},
    pages = {740--746},
    booktitle = {{IJCAI}},
    publisher = {{AAAI} Press},
    author = {Boyarski, Eli and Felner, Ariel and Stern, Roni and Sharon, Guni and Tolpin, David and Betzalel, Oded and Shimony, Eyal},
    year = {2015},
}

@article{andreychuk_improving_2021,
    title = {Improving Continuous-time Conflict Based Search},
    volume = {35},
    rights = {Copyright (c) 2021 Association for the Advancement of Artificial Intelligence},
    issn = {2374-3468},
    pages = {11220--11227},
    number = {13},
    journal = {{AAAI}},
    author = {Andreychuk, Anton and Yakovlev, Konstantin and Boyarski, Eli and Stern, Roni},
    year = {2021},
    langid = {english},
}

\vfill\pagebreak

\end{document}